\documentclass{article}
\usepackage{ijcai25}

\usepackage{times}
\usepackage{soul}
\usepackage{url}
\usepackage[hidelinks]{hyperref}
\usepackage[utf8]{inputenc}
\usepackage[small]{caption}
\usepackage{graphicx}
\usepackage{amsmath}
\usepackage{amsthm}
\usepackage{booktabs}
\usepackage{algorithm}
\usepackage{algorithmic}
\usepackage[switch]{lineno}
\usepackage{fontawesome5}
\usepackage{amsmath,amssymb,amsthm,mathtools,graphicx}
\usepackage{xcolor}
\usepackage{dsfont}

\newcommand{\corresp}{%
  \faIcon[regular]{envelope}%
}

\theoremstyle{plain}
\newtheorem{thm}{Theorem}

\newtheorem{prop}{Proposition}

\theoremstyle{definition}

\newcommand{\Eqref}[1]{Eq.~(\ref{#1})}

\newcommand{\Thmref}[1]{Theorem~\ref{#1}}

\newcommand{\Propref}[1]{Proposition~\ref{#1}}

\newcommand{\calX}{\mathcal{X}}
\newcommand{\calY}{\mathcal{Y}}

\newcommand{\calP}{\mathcal{P}}

\newcommand{\calS}{\mathcal{S}}

\newcommand{\R}{\mathbb{R}}

\newcommand{\lrpar}[1]{\left( #1 \right)}
\newcommand{\lrbra}[1]{\left[ #1 \right]}
\newcommand{\lrcubra}[1]{\left\{ #1 \right\}}

\makeatletter
\newcommand{\bigplus}{%
  \DOTSB\mathop{\mathpalette\mattos@bigplus\relax}\slimits@
}
\newcommand\mattos@bigplus[2]{%
  \vcenter{\hbox{%
    \sbox\z@{$#1\sum$}%
    \resizebox{!}{0.9\dimexpr\ht\z@+\dp\z@}{\raisebox{\depth}{$\m@th#1+$}}%
  }}%
  \vphantom{\sum}%
}
\makeatother

\DeclareFontFamily{U}{mathx}{}
\DeclareFontShape{U}{mathx}{m}{n}{<-> mathx10}{}
\DeclareSymbolFont{mathx}{U}{mathx}{m}{n}
\DeclareMathAccent{\widecheck}{0}{mathx}{"71}

\newcommand{\E}[1]{\mathbb{E}\left[ #1 \right]}
\newcommand{\V}[1]{\mathbb{V}\left( #1 \right)}
\newcommand{\pset}[1]{\calP_{#1}}

\newcommand{\eqdef}{:=}

\newcommand{\Cor}[1]{\textrm{Corr}\left( #1 \right)}
\newcommand{\Shap}{\textrm{Shap}}

\title{Beyond Shapley Values: Cooperative Games for the Interpretation of Machine Learning Models}

\author{
Marouane Il Idrissi$^{\textrm{\corresp}12}$
\and
Agathe Fernandes Machado$^1$
\and
Arthur Charpentier$^1$
\affiliations
$^1$Département de Mathématiques, Université du Québec à Montréal\\
$^2$Institut Intelligence et Données, Université Laval\\
\emails
\corresp ~ ilidrissi.m@gmail.com
}

\begin{document}

\maketitle

\begin{abstract}

Cooperative game theory has become a cornerstone of post-hoc interpretability in machine learning, largely through the use of Shapley values. Yet, despite their widespread adoption, Shapley-based methods often rest on axiomatic justifications whose relevance to feature attribution remains debatable. In this paper, we revisit cooperative game theory from an interpretability perspective and argue for a broader and more principled use of its tools. We highlight two general families of efficient allocations, the Weber and Harsanyi sets, that extend beyond Shapley values and offer richer interpretative flexibility. We present an accessible overview of these allocation schemes, clarify the distinction between value functions and aggregation rules, and introduce a three-step blueprint for constructing reliable and theoretically-grounded feature attributions. Our goal is to move beyond fixed axioms and provide the XAI community with a coherent framework to design attribution methods that are both meaningful and robust to shifting methodological trends.
\end{abstract}

\section{Introduction}
Over the past several years, methods derived from cooperative game theory have increasingly fueled the literature on XAI \cite{Barredo2020}. In particular, numerous local and global post-hoc feature attribution techniques are defined as cooperative game allocations. However, recent studies have highlighted that the lack of rigorous theoretical foundations for feature attributions significantly hinders the precise assessments of their correctness \cite{Haufe2024}. Common theoretical justifications rely primarily on axiomatic characterizations of the Shapley values \cite{Shapley1951}, but their interpretation as measures of feature importance remains ambiguous and has faced criticism \cite{Verdinelli2024}. Consequently, their practical reliability, especially in critical decision-making contexts, has been questioned \cite{Xin2024}. This conceptual gap makes it difficult to validate attribution methods beyond heuristic or empirical arguments.

In this paper, we provide a broader perspective on cooperative game theory specifically tailored to machine learning interpretability. Our contributions are twofold: i) we highlight the extensive range of possibilities available when looking beyond Shapley values, and ii) we demystify the mathematical foundations of cooperative game theory to enhance the theoretical understanding of attribution methods based upon it. Ultimately, we intend for this paper to serve as a blueprint for innovation in feature attribution, clarifying its theoretical intricacies to foster the development of robust attribution methods that withstand changing fashions in the field.

\subsection{Main contributions}
\paragraph{Richness of cooperative games for model interpretation.} We provide an accessible overview of cooperative game theory, specifically adapted for constructing post-hoc feature attributions. We introduce two general classes of allocations, the Weber and Harsanyi sets of allocations, that generalize the Shapley values and discuss their interpretative strengths. For each class, we revisit essential theoretical results from the existing literature, emphasizing their relevance for practical XAI applications. Additionally, we question the importance of axiomatic characterizations as robust theoretical justifications.

\paragraph{Recipe for game theoretic feature attributions.} We outline the key methodological steps involved in developing a game-theoretic feature attribution methods, emphasizing the interpretative consequences of different choices for value functions and allocation methods.  Using the Weber and Harsanyi sets, we pave the way towards creating custom and relevant allocation schemes, which can be tailored to any interpretability study. Furthermore, we highlight recent theoretical developments that offer rigorous guidance on these methodological decisions.

\subsection{Notations}
In the remainder of this article $d$ is a positive integer denoting the number of features, $X$ is the $d$-dimensional random vector of features valued in $\calX \subseteq \R^d$ and $Y$ represents the random target valued in $\calY \subseteq \R$. Denote by $f : \calX \rightarrow \calY $ a learned machine learning model whose goal is to predict $Y$ using $X$. Let $D = \{1,\dots,d\}$ and let $\pset{D}$ denote the set of subsets of $D$ (i.e., its powerset). For every $A \in \pset{D}$, let $X_A = \lrpar{X_i}_{i \in A}$, the $\calX_A \subseteq \R^{|A|}$ valued subset of features with indices in $A$. 


\section{Cooperative games and allocations}
\label{sec:coopGames}
A (transferable-utility) cooperative game is a tuple $(D,v)$ where $D = \{1,\dots,d\}$ is a set of players and  $v : \pset{D} \rightarrow \mathbb{R}$ is a value function, quantifying the value of all coalitions of players \cite{Osborne1994}. The dual of a cooperative game $(D,v)$ is the cooperative game $(D,w)$ where $w : A \mapsto v(D) - v(D\setminus A)$. An allocation of a cooperative game is a mapping $\phi_v : D \rightarrow \R$, attributing a payoff to all players individually. An allocation is efficient if $\sum_{j \in D} \phi_v(j) = v(D) - v(\emptyset)$, effectively redistributing the whole of $v(D)$ among the players. For example, the Shapley values of a cooperative game $(D,v)$ is the allocation, for every $j \in D$, defined as \cite{Shapley1951}
\begin{equation*}
    \Shap_v(j) = \frac{1}{d}\sum_{\underset{j \notin A}{A \in \mathcal{P}_D}} \binom{d-1}{|A|}^{-1} \left[v(A \cup \{j\}) - v(A)\right].
\end{equation*}
This particular allocation has gained significant traction in the XAI community in recent years. Its popularity originates from the analogy drawn between players in a cooperative game and features in a predictive model. However, the interpretation of the Shapley values in their original formulation can be somewhat unclear to practitioners. Consequently, the XAI community frequently relies on their axiomatic characterizations to justify their theoretical foundations, regardless of the specific application context. In the following subsections, we provide alternative interpretations of the Shapley values through the lenses of the random order and dividend-sharing paradigms.

\subsection{Weber set: Shapley values as the uniform distribution over permutations}
The paradigm underlying the Weber set is based on the notion of \emph{random orders}. Players are assumed to join a coalition sequentially, following a random ordering. For instance, consider $3$ players and an ordering $\pi =(2,3,1)$ occurring with probability $p(\pi)$.  Each player receives their \emph{weighted marginal contribution to the preceding players in this ordering}: player $2$ receives $p(\pi) \times \lrbra{v({2}) - v(\emptyset)}$, player $3$ receives $p(\pi) \times \lrbra{v({2,3}) - v({2})}$, and player $1$ receives $p(\pi) \times \lrbra{v({1,2,3}) - v({2,3})}$. Aggregating these contributions over all possible orderings yields an allocation. The Weber set is precisely the set of allocations expressible in this form, parameterized by a probability mass function over player orderings.

More formally, let $\calS_D$ be the set of permutations of $D$ (without replacement). For an ordering $\pi = (\pi_1, \dots, \pi_d) \in \calS_D$ and for any player $j\in D$, let $\pi(j)$ be the position of player $j$ in the ordering $\pi$ (i.e., $\pi_{\pi(j)}=j$), and let $\pi^j$ be the set of players preceding $j$ in $\pi$, including $j$ (i.e., $\pi^j \eqdef \lrcubra{i \in D: \pi(i) \leq \pi(j)}$). Let $p : \calS_D \rightarrow [0,1]$ be a probability mass function over $\calS_D$, called the random order distribution. The Weber set \cite{Weber1988} is the set of allocations, parametrized by $p$, that can be written, for every $j \in D$, as
\begin{equation}
     \phi_v(j) = \mathbb{E}_{p} \lrbra{v\lrpar{\pi^j} - v\lrpar{\pi^j \setminus \lrcubra{j}}}
    \label{eq:webAllocations}
\end{equation}
which can be interpreted as the expected marginal contribution of each player over all possible orderings drawn from $p$.
\begin{prop}[Efficiency of the Weber set]
    Allocations of the form of \Eqref{eq:webAllocations} are efficient.
    \label{prop:effWeber}
\end{prop}
We refer the interested reader to \cite{Weber1988} for a complete proof of \Propref{prop:effWeber}, and the appendix for a sketch of the proof.. Different choices of random order distributions can lead to different allocations.
\begin{thm}[Shapley and uniform orderings]
    For a cooperative game $(D,v)$, let $\phi_v$ be an allocation in the Weber set parametrize by the random order distribution $p$. Then, $\phi_v = \Shap_v$ if and only if $p(\pi) = 1/d!$ for every $\pi \in \calS_D$.
    \label{thm:shapWeber}
\end{thm}
A proof of \Thmref{thm:shapWeber} is available in \cite{Weber1988}. This result yields an equivalent formulation of the Shapley values, for every $j \in D$:
\begin{equation}
    \Shap_v(j) = \frac{1}{d!} \sum_{\pi \in \calS_D} \lrbra{v\lrpar{\pi^j} - v\lrpar{\pi^j \setminus \lrcubra{j}}}
    \label{eq:shapWeber}
\end{equation}

\subsection{Harsanyi set: Shapley values as the egalitarian redistribution of dividends}
The Harsanyi set of allocations relies on the concept of \emph{dividend redistribution}. Just as $v$ quantifies the cumulative worth of coalitions, the Harsanyi dividends $\varphi_v : \pset{D} \rightarrow \R$ quantify the incremental worth of a coalition of players. For instance, with $3$ players, we have $\varphi_v({j}) = v({j}) - v(\emptyset)$ for $j \in {1,2,3}$, representing the incremental value added by player $j$ alone. For coalitions of two players, the dividends are computed recursively: $\varphi_v({i,j}) = v({i,j}) - \varphi_v({i}) - \varphi_v({j}) - v(\emptyset)$. More generally, for any coalition $A \in \pset{D}$, the dividends are defined recursively as $\varphi_{v}(A) = v(A) - \sum_{B \subsetneq A} \varphi_v(B)$, equivalent to the explicit form given by \cite{Harsanyi1963}:
\begin{equation}
    \forall A \in \pset{D}, \enspace \varphi_v(A) = \sum_{B \in \pset{A}} (-1)^{|A|-|B|}v(B).
    \label{eq:harsDiv}
\end{equation}
Allocations in the Harsanyi set redistribute the dividends among the coalition members according to a weight system $\lambda : D \times \pset{D} \rightarrow \R$ satisfying, for every $j \in D$ and $A \in \pset{D}$, the conditions $\lambda_j(A) \geq 0$, $\sum_{j \in D} \lambda_j(A) = 1$, and $\lambda_j(A) = 0$ whenever $j \not \in A$. Formally, the allocations in the Harsanyi set are given by:
\begin{equation}
    \forall j \in D, \enspace \phi_v(j) = \sum_{A \in \pset{D}~ :~ j \in A} \lambda_j(A) \varphi_v(A).
    \label{eq:harsaAllocations}
\end{equation}

\begin{prop}[Efficiency of the Harsanyi set]
    Allocations of the form \Eqref{eq:harsaAllocations} are efficient.
    \label{prop:effHarsa}
\end{prop}
The efficiency of allocations in the Harsanyi set has been established in the literature. We refer the interested reader to \cite{Vasilev2001} for a complete proof of \Propref{prop:effHarsa}, and the appendix for a sketch of the proof. Different weighting systems yield different allocations.
\begin{thm}[Shapley and egalitarian redistribution]
    For a cooperative game $(D,v)$, let $\phi_v$ be an allocation in the Harsanyi set parametrized by the weight system $\lambda$. Then, $\phi_v = \Shap_v$ if and only if $\lambda_i(A) = 1/|A|$ for every $i \in D$, and $A \in \pset{D}$ such that $i \in A$.
    \label{thm:shapHarsa}
\end{thm}
\Thmref{thm:shapHarsa} offers another equivalent interpretation of Shapley values \cite{Harsanyi1963}: dividends from each coalition $A \in \pset{D}$ are equally shared among its members, leading to
\begin{equation}
    \forall j \in D, \enspace \Shap_v(j) = \sum_{A \in \pset{D} ~ : ~ j \in A} \frac{\varphi_v(A)}{|A|}.
    \label{eq:shapHarsa}
\end{equation}

\subsection{Efficiency is all you need}
Using the Weber and the Harsanyi set as references, one can see allocations as \emph{aggregations of coalitions' values}, either through random orders or dividends. The choice of a probability mass function over permutations or a weight system determines the intrinsic quality of this aggregation. However, it is important to emphasize that \emph{the aggregation process is distinct from the choice of value function}. This observation has two consequences: i) an allocation cannot correct a poorly chosen value function, and ii) the axioms typically used to justify feature attributions theoretically govern only the \emph{aggregation process}, not the choice of value function.

From this perspective, although groundbreaking in cooperative game theory, the axiomatic characterization of the Shapley values provides little theoretical justification in the context of XAI. Therefore, the critical property (or axiom) that an allocation must satisfy for XAI remains efficiency, as it directly relates the resulting attribution to the quantity of interest in the study. For example, in its original formulation, the SHAP feature attribution method \cite{Strumbelj2010,Lundberg2017}, for a sample $x$ of $X$, amounted to choosing the value function $v(A) = \E{f(X) \mid X_A=x_A}$ in conjunction with the Shapley values allocation. It can be argued that the main appeal of this method came from the fact that it decomposes a prediction $f(x)$, a property directly related to efficiency, and relies less on the choice of the aggregation process of the Shapley values.

Consequently, the most significant decision in feature attribution is \emph{the selection of the value function}. Choosing an allocation is akin to selecting a way of summarizing the information contained in $2^d$ coalition values into $d$ quantities. The following section will guide these choices and highlight promising future research directions.

\section{Leveraging cooperative games for model interpretation}
In this section, we introduce a three-step recipe to leverage cooperative game theory for model interpretation: 1) choosing a quantity of interest, 2) choosing a value function relevant to this quantity of interest, and 3) choosing an allocation for ease of interpretation. While the first step depends entirely on the specific XAI study, we present recent advances and perspectives on selecting suitable value functions and allocations.

\subsection{Overall blueprint}
The blueprint for constructing model-agnostic feature attributions to derive insights from machine learning models is straightforward and can be broken down into the following steps.

\paragraph{Step 1: Meaningful quantity of interest.}
The first step is to \emph{choose a quantity worth studying} \cite{Ilidrissi2023}. This choice can be guided by the nature of the XAI study or by the types of insights one aims to uncover. For an instance $x$ of $X$, examples of such quantities include the model's prediction $f(x)$ \cite{Strumbelj2010}, the model's variance $\V{f(X)}$ for uncertainty attribution \cite{Owen2014}, or more specialized model-related quantities, such as kernel-based or information-theoretic metrics \cite{DaVeiga2021,Watson2023}.

\paragraph{Step 2: Value function.}
The second step is to choose a value function that exploits the efficiency property of the subsequent allocations. In that regard, the essential requirement for the value function is that it equals the selected quantity of interest when evaluated on the full set of features, i.e., $v(D)$ equals the chosen quantity. As highlighted in \cite{Sundararajan2020}, multiple valid value functions can represent the same quantity of interest. For example, when decomposing predictions, both
$v(A) = \E{f(X) \mid X_A = x_A}$ and
$v'(A) = \int_{\R^{|D|-|A|}} f(x_{D\setminus A}, x_A) , dP_{X_{D \setminus A}}(x_{D\setminus A})$
satisfy $v(D) = v'(D) = f(x)$; hence, any efficient allocation based on these value functions decomposes the prediction. In sensitivity analysis, common value functions include
$v(A) = \V{\E{f(X) \mid X_A}}$ and
$v'(A) = \E{\V{f(X) \mid X_{D \setminus A}}}$ \cite{Song2016,Herin2024}.

\paragraph{Step 3: Allocation} Choosing an efficient allocation, while less crucial than selecting the value function, can nevertheless highlight interesting behaviors. For example, with variance decomposition and the standard value function, Shapley values can provide insightful interpretations when features are independent \cite{Iooss2019}, and proportional values \cite{Ortmann2000} are helpful in detecting spurious features \cite{Herin2024}. Leveraging efficiency ensures a clear interpretation: the quantity of interest is fully redistributed among the features.

\subsection{Value functions and model representations}
The choice of value function can lead to legitimate or misleading insights into the behavior of black-box models. Although several empirical studies have compared and proposed value functions \cite{Saumendu2022,Amoukou2022}, only recent work has introduced theoretical criteria for choosing appropriate value functions. To illustrate these recent advancements, let $d=2$, $X_1, X_2 \sim \mathcal{N}(0,1)$ with $\Cor{X_1, X_2} = \rho$. Let $x=(x_1, x_2)$, select the task of prediction decomposition, with value function $v(A) = \E{f(X) \mid X_A = x_A}$, and let the simple model $f(x) = x_1 + x_2 + x_1x_2$. One can analytically compute the resulting Shapley values:
$$\Shap_v(\{1\}) = x_1 + \frac{\rho}{2}(x_1 +x_1^2 -x_2 -x_2^2-1) +\frac{x_1x_2}{2},$$
$$\Shap_v(\{2\}) = x_2 + \frac{\rho}{2}(x_2 +x_2^2 -x_1 -x_1^2-1) +\frac{x_1x_2}{2}.$$
While this decomposition correctly redistributes the model output, it arguably lacks fidelity to the original analytical expression. Although not formally defined, this fidelity gap has been referred to as a lack of \emph{purity} in the literature \cite{Kohler2024}.

Recent developments in sensitivity analysis and high-dimensional model representation offer systematic methods for choosing value functions that yield pure allocations. These approaches replace conditional expectations, which are orthogonal projections, with oblique projections onto the same subspaces, explicitly incorporating feature dependencies. Revisiting our previous example with these oblique-projection-based value functions, \cite{IlIdrissi2025} derived the following Shapley values:
$$\Shap_v(\{1\}) = x_1  +\frac{x_1x_2}{2}, \enspace \Shap_v(\{2\}) = x_2 + \frac{x_1x_2}{2}.$$
This representation is arguably more faithful to the model's analytical structure. These theoretical advancements indicate that theoretically grounded choices of value functions remain central to cooperative game-based feature attribution. Another critical insight from these developments is that \emph{the problem of feature attribution is a problem of representation}. Different methods decompose the quantity of interest differently, and the absence of ground truth limits empirical comparisons. Therefore, theoretical results of this nature are essential for providing robust justifications and guidance on the choice of method.

\subsection{Purposeful allocations}
The vast majority of methods in the literature on machine learning interpretability rely on well-established allocations. While these allocations are meaningful within cooperative game theory, their relevance for interpretability tasks is questionable. Although the analogy between cooperative game players and features initially appears intuitive, several core concepts from cooperative game theory may be irrelevant or misleading for interpreting machine learning models. By leveraging the Weber and Harsanyi sets, however, we can formalize new classes of allocations tailored to specific XAI needs.

This has already been successfully done in the literature. For instance, \cite{Herin2024} introduced the proportional marginal effects (PME) for the task of variance decomposition, with $v(A) = \V{\E{f(X) \mid X_A}}/\V{f(X)}$. They adapted the \emph{proportional values} \cite{Ortmann2000} to the dual of the cooperative game, yielding theoretical guarantees on the detection of spurious features when mutual independence does not hold. Moreover, through various studies on real-world datasets, it has been shown that the PME yields more accurate importance quantification than the Shapley values in challenging situations (highly correlated features). In a simple analytical example where
$$ X = \begin{pmatrix}X_1 \\ X_2 \\ X_3 \end{pmatrix} \sim \mathcal{N} \lrpar{\begin{pmatrix} 0\\0\\0\end{pmatrix}, \begin{pmatrix}1 &0 &\rho \\ 0 & 1 & 0 \\ \rho & 0 & 1\end{pmatrix}} \enspace,$$
with model $f(X) = X_1 + X_2$, where $X_3$ is spurious by design. The Shapley values and PMEs can be computed analytically yielding
\begin{align*}
    &\Shap_w(\{1\})= 0.5(1-\rho^2/2), \enspace\Shap_w(\{2\}) = 0.5, \\
    &\Shap_w(\{3\})= \rho^2/4,\\
    &\textrm{PME}_w(\{1\})=0.5, \enspace \textrm{PME}_w(\{2\})=0.5, \enspace \textrm{PME}_w(\{3\})=0.
\end{align*}
In this illustration, the traditional Shapley values are sensitive to $\rho$, while the PME are not. Moreover, the PMEs do not grant importance to the spurious feature $X_3$ while the Shapley values do as soon as mutual independence is not respected.

More generally, through the lens of the Weber set, choosing an allocation is equivalent to selecting a probability distribution over permutations of $D$. In this framework, Shapley values, corresponding to the uniform distribution, represent the maximum-entropy distribution. By altering optimization objectives or incorporating constraints, one can define new optimal distributions and, thus, novel allocations. For instance, \cite{Frye2020} proposed to modify the random order distribution to echo some a-priori causal structure between the features.

From the Harsanyi dividends perspective, the weighting system can be interpreted as a sparse row-stochastic matrix of dimension $2^d \times d$. While it is unclear how to interpret Shapley values in this setting meaningfully, optimal transportation theory could be employed to derive new weighting systems, and thus new allocations optimized for various interpretability-driven costs. Such approaches have yet to be explored in the literature, despite their potential to enhance theoretical justifications in game-theoretic interpretability methods.

\section{Conclusion}
In this paper, we have revisited cooperative game theory as a powerful framework for constructing meaningful feature attributions beyond the widely used Shapley values. We highlighted two general allocation families and demonstrated their expressiveness, interpretative flexibility, and theoretical richness. Aside from the efficiency property, the relevance of traditional axioms in justifying the choice of allocation for XAI purposes is discussed. Moreover, by clarifying the separation between the choice of allocation and the selection of the value function, we provided a systematic 3-step blueprint to guide practitioners in their interpretation endeavors.

We hope these perspectives inspire the development of reliable, interpretable tools that are resistant to changing trends, thus significantly enriching the toolbox available for XAI.

\appendix
\section{Proofs}
\label{sec:proofs}

\begin{proof}[Sketch for the proof of Proposition~\ref{prop:effWeber}]
    Notice that, for any $\pi \in \calS_D$, $\sum_{j \in \pset{D}} \lrbra{v\lrpar{\pi^j} - v\lrpar{\pi^j \setminus \lrcubra{j}}}$ is a telescoping series, which is equal to $v(D) - v(\emptyset)$. The result comes from the fact that $p$ is a pmf.
\end{proof}

\begin{proof}[Sketch for the proof of Proposition~\ref{prop:effHarsa}]
    Using Rota's generalization of the Möbius inversion formula on powersets \cite{Rota1964}, notice that $v(D) = \sum_{A \in \pset{D}} \varphi_v(A)$. Then, efficiency follows from the fact that the sum of the weights over the players in valid weight systems equals one.
\end{proof}

\bibliographystyle{named}
\bibliography{paper}

\end{document}